\documentclass[11pt,a4paper]{article}
\usepackage[hyperref]{emnlp2018}
\usepackage{times,graphicx}
\usepackage{latexsym,booktabs,dingbat}
\usepackage{tikz-dependency}
\usepackage{url}
\usepackage[]{todonotes}   % insert [disable] to disable all notes

\aclfinalcopy % Uncomment this line for the final submission
\usepackage{array, booktabs, adjustbox, multirow}
\usepackage{blindtext}

\title{Nightmare at test time:\\How punctuation prevents parsers from generalizing}

\author{Anders S\o gaard$^{1}$
~~~Miryam de Lhoneux$^{2}$~~~Isabelle Augenstein$^{1}$\\[1mm] 
\begin{tabular}{cp{10mm}c}
    $^{1}$Dpt.~of Computer Science && $^{2}$ Dpt.~ of Linguistics and Philology \\ 
    University of Copenhagen && Uppsala University \\ 
\end{tabular}
}

\date{}

\begin{document}
\maketitle
\begin{abstract}
Punctuation is a strong indicator of syntactic structure, and parsers trained on text with punctuation often rely heavily on this signal. Punctuation is a diversion, however, since human language processing does not rely on punctuation to the same extent, and in informal texts, we therefore often leave out punctuation. We also use punctuation ungrammatically for emphatic or creative purposes, or simply by mistake. We show that (a) dependency parsers are sensitive to {\em both} absence of punctuation and to alternative uses; (b) neural parsers tend to be more sensitive than vintage parsers; (c) training neural parsers {\em without}~punctuation outperforms all out-of-the-box parsers across all scenarios where punctuation departs from standard  punctuation. Our main experiments are on synthetically corrupted data to study the effect of punctuation in isolation and avoid potential confounds, but we also show effects on out-of-domain data.
\end{abstract}

\section{Introduction}

\noindent We study the sensitivity of modern dependency parsers to punctuation. While punctuation was originally motivated by reading aloud, serving the purpose of ``breath marks'' \cite{Baldwin:Coady:78}, many modern-day punctuation systems are designed to facilitate grammatical disambiguation. 
This paper aims to show that for this reason, punctuation can significantly hurt the generalization ability of state-of-the-art syntactic parsers. In other words, syntactic parsers become too reliant on punctuation and therefore suffer from the absence or creative uses of punctuation. Such uses are abundant; see Table~1 for examples from Twitter. Such situations, where highly predictive features are absent or distorted at test time, were referred to in \newcite{Globerson:Roweis:06} as {\em nightmare at test time}.
Human reading is very robust to variation in punctuation \cite{Baldwin:Coady:78}; so creative use of punctuation does not hurt human reading performance. In effect, sensitivity to punctuation is a major obstacle that prevents our syntactic parser from 
achieving human-level robustness. 

\begin{table}[tp]
\fontsize{9}{9}\selectfont
\begin{center}
\begin{tabular}{ll}
\toprule&{\bf No punctuation}\\[4pt]
(1)&i have so many questions i dont know where to start\\[4pt] 
\midrule&{\bf Creative punctuation}\\[4pt]
(2)&
What. The. Fuck. Ever. Dot. Com\\[4pt]
(3)&
\ldots and then , , , , i start to feel $\sim$lonely$\sim$\\[4pt]
\midrule&{\bf Both}\\[4pt]
(4)&
I feel like ... idk ... idk ... idk man. Nvm I'm good. \\[4pt]

\bottomrule
\end{tabular}
\end{center}
\caption{Examples of uses of punctuation}\label{tab:examples}
\end{table}

The generalization ability of a dependency parser is usually measured by evaluating its accuracy on held-out data,  our yardstick to prevent over-fitting, i.e.~we define the degree to which a parser has over-fitted to the training data as the difference between performance on training data and performance on the held-out data. This practice is poor when data is not i.i.d., since the held-out data cannot be assumed to be representative;  
in such cases, little or no over-fitting does not guarantee our parsers have learned important linguistic generalizations: Rather, the parsers may have over-fitted to superficial cues that are  present in {\em both} the training and test datasets
\cite{Jo:Bengio:17}. 
We argue that punctuation signs are superficial cues preventing modern parsers from learning appropriately high-level abstractions from our datasets. 

\paragraph{Contributions} We evaluate three neural dependency parsers for English, as well as two older alternatives, on a standard benchmark, before and after stripping punctuation, as well as after injecting more punctuation  
signs in the benchmark. We show that (a) projective parsers are, unsurprisingly, more sensitive to punctuation injection than non-projective ones, since punctuation injection may introduce crossing edges, and (b) neural parsers are more sensitive than vintage parsers. The latter is our main contribution, but we also show that training a neural parser {\em without}~punctuation outperforms all parsers trained in a regular fashion 
across all punctuation scenarios. Our experiments are on semi-synthetic data to control for confounds, but we also show the parser trained without punctuation is superior on real data with non-standard punctuation. 

\begin{figure}
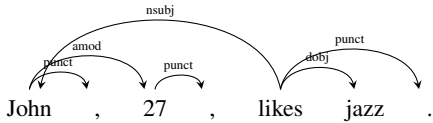

\begin{center}
\begin{dependency}[theme = simple, edge style={<-}]
\begin{deptext}[column sep=1em]
{\small John} \& {\small ,} \& {\small 27} \& {\small ,} \& {\small likes} \& {\small jazz} \& {\small .}\\
\end{deptext}
\depedge{1}{5}{nsubj}
\depedge{2}{1}{punct}
\depedge{3}{1}{amod}
\depedge{4}{3}{punct}
\depedge{6}{5}{dobj}
\depedge{7}{5}{punct}
\end{dependency}
\end{center}
\caption{\label{tree}Punctuation in Stanford dependencies}
\end{figure}

\section{Punctuation in Stanford dependencies}

\paragraph{Dependency annotation} Dependency annotation refers to the manual assignment of syntactic structures to sentences, following one of several sets of available annotation guidelines. This paper focuses exclusively on the Stanford dependencies annotation scheme \cite{Marneffe:Manning:08}. This scheme restricts the set of possible syntactic structures to single-rooted, ordered, possibly non-projective trees whose edges are uniquely labeled by a single dependency label.

\paragraph{Punctuation} Punctuation should be distinguished from diacritics and logographs. The two most frequently used punctuation signs are periods and commas. Periods (``.''), however, are potentially ambiguous with other uses of dots, typically indicating omissions or pauses. When dots are used emphatically and creatively it is hard to maintain this distinction, and we will simply refer to {\em dots} and {\em commas} in this paper. We ignore other punctuation signs, including dashes, question and exclamation marks, and colons and semicolons. 

Punctuation is, among other things, used to mark boundaries between constituents of written language. Space characters, for example, separate words, albeit sometimes inconsistently. Spacing is a fairly recent innovation in writing; classical Latin and Greek did not leave spaces between words, and many Asian languages, e.g., Thai and Lao, still do not.  
A period is typically used to mark the end of a grammatical sentence, and commas are often used to separate clauses. Therefore, punctuation also correlates strongly with properties of syntactic structures and is therefore very predictive of dependency structures. 

Variation in punctuation is often observed in informal texts, but variation may also be the result of errors. Punctuation errors are by far the most frequent error type in scientific writing, for example \cite{Remse:ea:16}. Modern parsers should be robust to such variation, just like humans are \cite{Baldwin:Coady:78}. 

\paragraph{Punctuation in Stanford dependencies} In the Stanford dependencies \cite{Marneffe:Manning:08}, periods attach to root tokens, and commas attach to their left neighbor or to root tokens; see Figure~\ref{tree}. 

\section{Experiments}

This section describes how we remove and inject punctuation (our perturbation maps), and details of the parsers used in our experiments. 

\paragraph{Perturbation maps} Since dots consistently attach to the root token of the sentence, and commas attach to their left neighbour or to the root token, we can remove and inject additional punctuation in a sentence without affecting the rest of its syntactic structure and without violating the wellformedness of dependency trees. Note, however, that injecting a root-dominated dot or comma may lead to crossing edges, i.e., turn a projective dependency tree into a non-projective one. This may lead to cascading errors for projective dependency parsers \cite{Ng:Curran:15}. In our experiments, arc-eager {\sc MaltParser} and {\sc Stanford} are the only projective parsers.  
We therefore propose two {\em perturbation maps} \cite{Jo:Bengio:17}: (a) simply removing punctuation, and (b) a simple injection scheme with two parameters $\chi$ and $\delta$. Let a dependency structure be an ordered tree with $n$ nodes decorated with words $w_1,\ldots,w_n$. At any node $1\leq i \leq n$, we (a) inject a comma at position $i$ with probability $\chi$ and move nodes $i\leq j\leq n$ to positions $j+1$, increasing the size of the graph by 1; and (b) inject a dot at position $i+1$ with probability $\delta$ and move nodes $i<j\leq n$ to positions $j+1$, increasing the size of the graph by 1. If we follow standard methodology and ignore punctuation when evaluating parsers, we can compare evaluations before and after applying the injection scheme. It is equally straightforward to remove punctuation without affecting the rest of the dependency tree. Each element $w_i$ to the right of punctuation nodes $w_j$ ($i>j$) moves to the left ($j-1$) for every punctuation item, decreasing the length of the sentence by 1 each time.

Note that both removing punctuation and our injection scheme can be seen as perturbation maps \cite{Jo:Bengio:17} of our dataset, with the following important properties: (a) grammatical structure recognizability, i.e., human ability to correctly process sentences, is preserved \cite{Baldwin:Coady:78}, (b) surface statistical regularities are qualitatively different,  
and (c) there exists a non-trivial generalization map between the original dataset and the perturbed version. These properties mean we can use our punctuation injection scheme to evaluate the sensitivity of neural dependency parsers to the surface statistical regularities involving dots and commas \cite{Jo:Bengio:17}. Since human reading is largely unaffected by erroneous punctuation, we may expect parsers to be robust to absence of punctuation and punctuation injection, as well. Our results clearly show this is not the case; in fact, recently proposed neural dependency parsers are {\em very} sensitive to differences in punctuation. 

\begin{table}
\begin{center}\begin{footnotesize}
\begin{tabular}{l|ccc}
\toprule
{\bf Parser}&{\bf Neural}&{\bf Trans.-based}&{\bf Projective}\\
\midrule
{\sc UUParser}&\checkmark&\checkmark&\\
{\sc KGraphs}&\checkmark&&\\
{\sc MaltParser}&&\checkmark&\checkmark\\
{\sc TurboParser}&&&\\
{\sc Stanford}&\checkmark&\checkmark&\checkmark\\
\bottomrule
\end{tabular}\end{footnotesize}
\end{center}
\caption{Our dependency parsers}
\end{table}

\paragraph{Our dependency parsers} We use five parsers in our experiments: the Uppsala parser ({\sc UUParser}) \cite{delhoneux17raw,delhoneux17arc}, the graph-based parser proposed in \cite{kiperwasser16}({\sc KGraphs}) , the arc-eager {\sc MaltParser} \cite{nivre07maltparser}, the {\sc TurboParser} \cite{fernandez15turbo}, and the {\sc Stanford} parser \cite{chen14fast}. {\sc UUParser} is a neural transition-based dependency parser, while {\sc KGraphs} is a neural graph-based parser. {\sc MaltParser} is a more traditional transition-based parser, and {\sc TurboParser} is a more traditional graph-based parser. Finally, the {\sc Stanford} parser is a projective, neural transition-based dependency parser. All parsers rely on predicted part-of-speech tags, except {\sc UUParser} (which does not rely on part-of-speech information at all). We use the {\sc TurboTagger} to obtain those. See Table~2 for an overview of our parsers.  

Finally, we also evaluate three non-standard versions of the {\sc UUParser}, namely, a parser trained with the same parameters as the off-the-shelf parser \cite{delhoneux17arc}, but which simply {\em ignores}~dots and commas completely ({\sc NoPunct}), and two heavily regularised versions of the parser trained in the standard fashion: (a) a version trained with the drop-out parameter set to 0.8 (zeros out 80\%~of activations); (b) a version with the gradient clipping parameter set to 0.075. We do so to answer the question of whether more heavily regularized dependency parsers are less sensitive to punctuation (they are not).

\begin{table*}
\begin{center}
\resizebox{\textwidth}{!}{%
\begin{tabular}{l|c|c|c|ccccc|c|cc|cc}
\toprule
\multicolumn{10}{c}{\sc English Penn Treebank (corrupted)}&\multicolumn{4}{c}{\sc Out-of-domain}\\
\midrule
&$\delta$=0&{\sc no}&Rel.err.&$\delta$=0.01&$\delta$=0.01&$\delta$=0.05&$\delta$=0.05&$\delta$=0.1&Rel.err.&\multicolumn{2}{c}{\sc Gweb}&\multicolumn{2}{c}{\sc Foster}\\
&$\chi$=0&{\sc punct}&incr.&$\chi$=0.01&$\chi$=0.05&$\chi$=0.01&$\chi$=0.05&$\chi$=0.1&incr.&{\sc Answ}&{\sc Rev}&{\sc Football}&{\sc Twitter}\\
\midrule
\midrule
{\sc UUParser}&{\bf 0.918}&0.869&0.598&{\bf 0.901}&0.867&0.886&0.851&0.794&1.512&{\bf 0.676}&0.662&0.770&0.699\\
{\sc KGraphs}&0.910&0.865&0.500&0.894&0.861&0.876&0.841&0.779&1.456&0.645&0.609&0.774&{0.715}\\
{\sc MaltParser}&0.858&0.805&0.373&0.836&0.791&0.804&0.757&0.675&1.289&0.605&0.566&0.721&0.642\\
{\sc TurboParser}&0.894&0.852&0.396&0.883&0.858&0.875&0.851&0.802&0.868&0.640&0.595&0.766&{\bf 0.722}\\
{\sc Stanford}&0.870&0.816&0.415& 0.845&0.808&0.806&0.772&0.688&1.400&0.640&0.608&0.735&0.689\\
\midrule
            {\sc no punct}&0.898&{\bf 0.898}&0.000&0.898&\multicolumn{4}{c}{\bf 0.898}&0.000&{0.670}&{\bf 0.669}&{\bf 0.792}& 0.701\\
    \midrule        
    {\sc dropout $\alpha$=0.8 }&0.904&0.847&0.594&0.884&0.845&0.858&0.820&0.748&1.625&0.661&0.652&0.761&0.682\\
            {\sc clip $t$=0.075}&0.917&0.871&0.554&0.900&0.864&0.887&0.851&0.793&1.494&0.672&0.657&{\bf 0.792}&0.676\\
            \bottomrule

\end{tabular}}
\caption{Labeled attachment scores with punctuation removed. 
 All parsers suffer from absence of or additional punctuation.  
 The relative increase in error ($\frac{\mbox{\sc 1-bl}}{\mbox{\sc 1-sys}}-1$; with {\sc bl} performance on original text; {\sc sys} performance under {\sc no punct} and $\delta=0.1,\kappa=0.1$, resp.) for neural parsers is higher than for non-neural parsers. {\sc Gweb} and {\sc Foster}~scores are on  development sentences (of at least five words) with no punctuation.}
\end{center}
\end{table*}

\section{Results and analysis}

We discuss the sensitivity of off-the-shelf dependency parsers to our perturbation maps, comparing to a parser trained after removing punctuation in the training data, as well as to heavily regularised versions of the same parser.

\paragraph{No punctuation} We first test our parsers on a version of the validation set where we strip away all punctuation. The data thus consists of newswire (WSJ 22) with punctuation removed. This is similar to Example (1) in Table~1, but in-domain. The results are in the second results column in Table~3, with the relative increases in error listed in the third results column. The drop induced by removing punctuation is quite dramatic: The {\sc UUParser}, for example, suffers an absolute drop of 5.4\%~LAS or an error increase of 67\%. For every three mistakes, {\sc UUParser} does, stripping away punctuation makes it introduce another two. 
Note that, generally, the relative increase in error is much higher for the three neural parsers, and that the regularisation strategies (drop-out and gradient clipping) do not seem to help much. 

\paragraph{Comma and dot injection} 
At medium injection rates, all parsers are  sensitive to punctuation injection. With $\delta=0.05,\gamma=0.05$, for example, all parsers perform worse than in the absence of punctuation. Our main observation is, again, that neural parsers suffer higher relative increases in errors than vintage parsers. Note that the {\sc MaltParser} is a projective parser and therefore has a higher relative increase in error; but {\sc TurboParser} is much more robust than the other parsers. That said, it still does much worse than the {\sc UUParser} trained without punctuation. 

\paragraph{Evaluation on informal text with non-standard punctuation} We also evaluate the models on sentences with non-standard punctuation in the development sections in the Google Web Treebank with informal text (from Yahoo Answers and user reviews). Specifically, we evaluate the models on sentences with more than one dot. Again, we show that the neural dependency parser trained without punctuation is superior to the other parsers.  

\section{Related work}

\paragraph{Punctuation in parsing}

\newcite{Spitkovsky:ea:11}~introduced the idea of splitting sentences at punctuation and
imposing parsing restrictions over the fragments and observed significant improvements in the context of unsupervised parsing. \newcite{Ng:Curran:15} aim to prevent cascading errors by enforcing correct punctuation arcs. They  restrict themselves to projective dependency parsing; erroneous punctuation arcs do not lead to cascading errors in non-projective dependency parsing. \newcite{Ma:ea:14}, motivated by the same observation
, treat punctuation marks as properties of their neighboring words rather than as individual tokens, showing improvements on in-domain data.

\paragraph{Breaking NLP models} 
\citet{jia-liang:2017:EMNLP2017} show how machine reading models can easily be broken with distractor sentences at test time and propose an alternative evaluation scheme, and \citet{journals/corr/abs-1711-02173} show how susceptible character-based machine translation models are to noise. Both papers are similar to ours in evaluating the performance of state-of-the-art models under corruptions of the data. There was recently a workshop dedicated to evaluation of NLP models under human adversarial example selection \cite{ettinger-EtAl:2017:BLGNLP2017}. Historically, NLP models were rarely evaluated on synthetic or otherwise adversarial data, but we believe this is a fruitful research direction. This is largely a philosophical question, and we believe a philosophical argument is in order. John Dewey \cite{Dewey:10}, the American philosopher, distinguishes three modes of thinking: (i) common reasoning, which identifies pattern in available, historical data, (ii) empirical thinking, which  collects new data to vary the experimental conditions, and (iii) experimental thinking, which actively modifies the conditions in controlled experiments to isolate the relevant variables. We believe recent work on breaking NLP models is an attempt to introduce experimental thinking into NLP, which has otherwise been limited -- or {\em handicapped} in Dewey's words -- by what  data happens to be available.

\section{Conclusions}

We evaluate the sensitivity of five dependency parsers to variations in punctuation, showing that available neural parsers tend to be more sensitive to such variation. We also show, however, that training neural parsers without punctuation provides a robust model that is better than any off-the-shelf parsers.

\section*{Acknowledgments}
We thank CSC in Helsinki and Sigma2 in Oslo for providing the computational resources used in the experiments, through NeIC-NLPL (www.nlpl.eu). The first author was supported by an ERC Starting Grant. 
\bibliography{acl2018}
\bibliographystyle{acl_natbib}

\end{document}